\documentclass{article}
\usepackage{arxiv}
\usepackage{microtype}
\usepackage{subcaption}
\usepackage{graphicx}
\usepackage{booktabs} % for professional tables
\usepackage{xspace}

\usepackage{amsmath}
\usepackage{amsfonts}
\usepackage{amsthm}
\usepackage{xargs} %allows /newcommandx
\usepackage{xparse}
\usepackage{mathtools}

\usepackage{hyperref}
\usepackage{cleveref}

\usepackage{breakurl}

\newcommand{\topk}{\textsc{Top-K}\xspace}

\newcommand{\sigmoid}{\textsc{Sigmoid}\xspace}
\newcommand{\softmax}{\textsc{Softmax}\xspace}

\newcommand{\acosp}{\textsc{ACoSP}\xspace}

\newcommand{\camvid}{\textsc{CamVid}\xspace}
\newcommand{\city}{\textsc{Cityscapes}\xspace}
\newcommand{\voc}{\textsc{Pascal VOC2012}\xspace}
\newcommand{\ade}{\textsc{Ade20k}\xspace}
\newcommand{\imagenet}{\textsc{ImageNet}\xspace}
\newcommand{\cifar}{\textsc{Cifar10}\xspace}
\newcommand{\segnet}{SegNet\xspace}
\newcommand{\pspnet}{PSPNet\xspace}

\newcommand{\resneteight}{ResNet-18\xspace}
\newcommand{\resnetfive}{ResNet-50\xspace}
\newcommand{\vgg}{VGG-16\xspace}

\newcommand{\sgd}{SGD\xspace}
\newcommand{\adam}{Adam\xspace}

\newcommand{\percdown}{\%\downarrow}
\newcommand{\github}{https://github.com/merantix/acosp}

\definecolor{codingbg}{rgb}{0.95,0.95,0.95}

\newcommand{\yrcite}[1]{\citeyearpar{#1}}
\renewcommand{\cite}[1]{\citep{#1}}

\newcommand\shorttitle{\textsc{ACoSP}}
\newcommand\authors{Ditschuneit \& Otterbach}

\title{Auto-Compressing Subset Pruning for Semantic Image Segmentation}
\date{}

\author{
Konstantin Ditschuneit and Johannes S. Otterbach \\
Merantix Labs GmbH, Berlin, Germany \\
\texttt{\{konstantin.ditschuneit, johannes.otterbach\}@merantix.com}
}

\begin{document}
\maketitle
\begin{abstract}
    State-of-the-art semantic segmentation models are characterized by high parameter counts and slow inference times, making them unsuitable for deployment in resource-constrained environments. To address this challenge, we propose \textsc{Auto-Compressing Subset Pruning}, \acosp, as a new online compression method. The core of \acosp consists of learning a channel selection mechanism for individual channels of each convolution in the segmentation model based on an effective temperature annealing schedule. We show a crucial interplay between providing a high-capacity model at the beginning of  training and the compression pressure forcing the model to compress concepts into retained channels. We apply \acosp to \segnet and \pspnet architectures and show its success when trained on the \camvid, \city, \voc, and \ade datasets. The results are competitive with existing baselines for compression of segmentation models at low compression ratios and outperform them significantly at high compression ratios, yielding acceptable results even when removing more than $93\%$ of the parameters. In addition, \acosp is conceptually simple, easy to implement, and can readily be generalized to other data modalities, tasks, and architectures.
    Our code is available at \url{\github}.
\end{abstract}

\section{Introduction}
\begin{figure}[t]
    \centering
    \includegraphics[width=.55\linewidth]{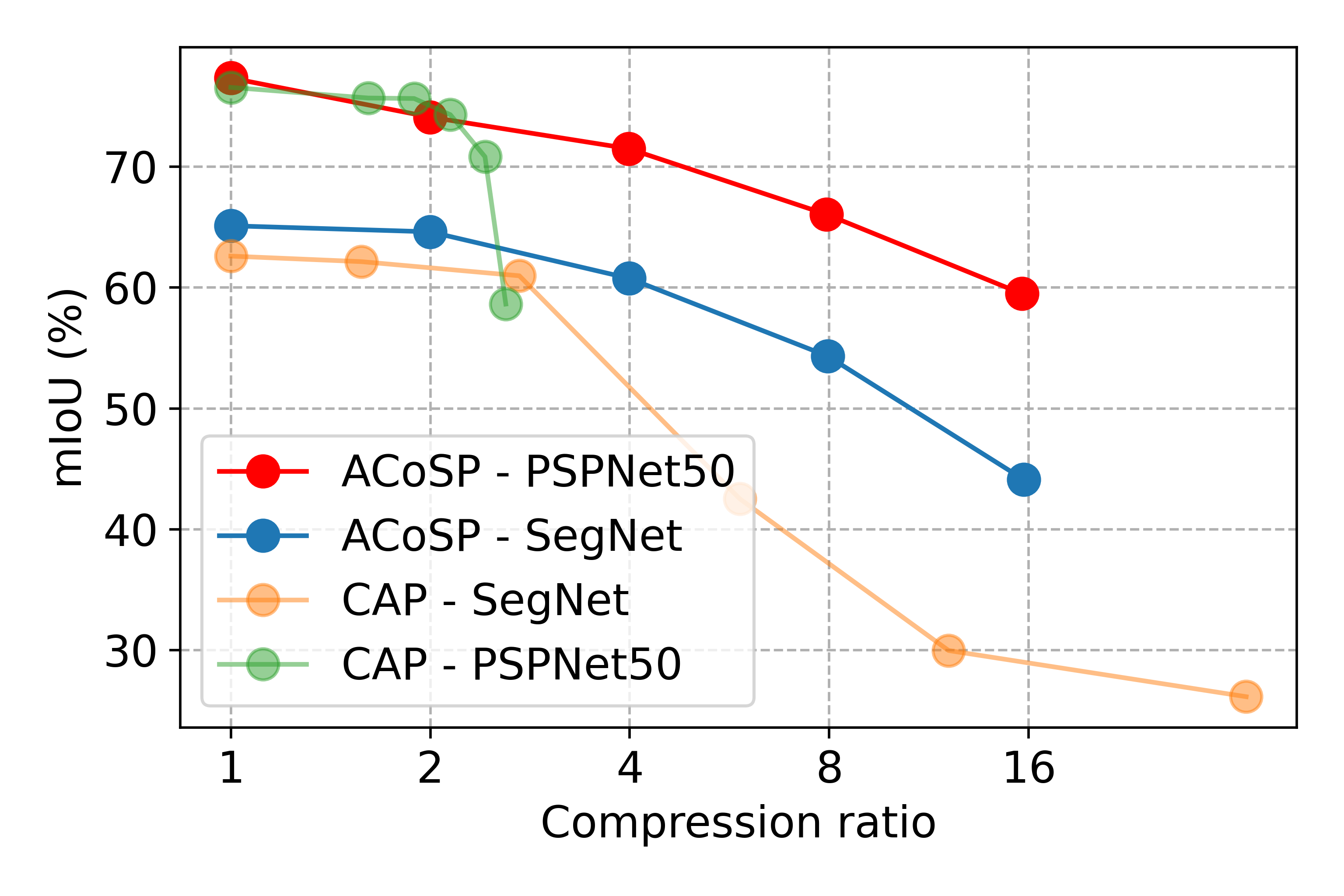}
    \caption{Mean Intersection-over-Union (mIoU) as a function of the compression ratio for \acosp and the baseline \textsc{CAP}~\cite{he_cap_nodate}. The data is obtained by training \segnet and \pspnet models on the \city dataset. For training details, see \Cref{sec:experiments} and \Cref{sec:implementation_details}. We can clearly see that we are competitive with \textsc{CAP} at small compression ratios and significantly outperform it at high compression ratios. Note that we do not have access to the trained baseline model and the reported numbers of parameters for their \pspnet differs from the official \textsc{PyTorch} implementation \cite{semseg2019}.}
    \label{fig:pruning_ratios}%
\end{figure}
Recent years have seen significant advances in training convolutional neural networks (CNNs), enabling new applications in semantic segmentation. However, these advances come at the cost of high computational resource requirements. State-of-the-art models typically have on the order of $10^7$ parameters, rendering a deployment in environments with a tight computational budget, such as autonomous vehicles or on-edge devices, challenging.

To reduce the computational burden of existing state-of-the-art models and maintain a high performance across tasks, the community is actively researching methods to efficiently compress and speed up neural networks. Pruning as a compression technique has been studied in great depth for image classification tasks \cite{Blalock2020}, but it has received less attention in the context of semantic segmentation. Latter, to the best of our knowledge, has only been discussed in a few prior works \cite{he_cap_nodate, you_gate_2019, bejnordi_batch-shaping_2020}.

In this work, we focus on model compression for semantic image segmentation using CNNs. We introduce a new pruning approach called \textsc{Auto-Compressing Subset Pruning}, \acosp for short, based on a modification of \textit{Differential Subset Pruning} which was first introduced in Li et al.~\yrcite{li_differentiable_2021}. \acosp efficiently reduces the number of convolution filters in the pruned model, resulting in a \textit{thin} network compared to the unpruned counterpart. While most pruning approaches  focus on finding the parameters or filters that can safely be removed, \acosp focuses on the removal process itself. Moreover, the chip design of current accelerator hardware, such as GPUs or TPUs, limits their ability to efficiently deal with arbitrarily and sparsely connected networks resulting from unstructured pruning approaches, resulting in significant engineering efforts for efficient deployment~\cite{Gajurel2021GPUAO, Zhang2016CambriconXAA}. Hence, the goal of \acosp is to combine the parameter efficiency of unstructured pruning while preserving a structured architecture to ease the model deployment on current hardware. In short, \acosp produces a thin, densely connected, and structured model.

To this end, we leverage that higher parameter counts correlate with a decrease in the number of training steps required to train a model to a certain performance, as studied in transformer-based models~\cite{kaplan_scaling_2020, strudel2021} and image classification~\cite{Rosenfeld2020ACP}. At the same time, the \textit{Lucky Lottery Hypothesis}~\cite{frankle_lottery_2019} tells us that only a few model parameters are relevant for training a model successfully, leading to sparsely connected network graphs. The key insight is to let the model use all of the parameters at the beginning of the training and then gradually decrease the importance of most convolution filters during training. In this way, we incentivize the model to compress most of the relevant parameters into the non-pruned layers while leveraging the lucky initialization and sample efficiency of large models at the beginning of the training.

With this approach, we turn pruning into a form of continuous auto-compression during the training, thus combining the benefits of high parameter counts during training with thin, efficient networks at inference time. In contrast to other learned selection approaches, our approach offers a simple selection process with minimal overhead during training, a strictly enforced pruning rate, and no additional loss function. This aids deployment in resource-constrained environments and preserves the guarantees of the original objective function as no additional losses need to be optimized during the compression process. To summarize our contributions:

\begin{itemize}
    \item We introduce a new pruning approach for image segmentation based on Differential Subset Pruning~\cite{li_differentiable_2021} that gradually compresses the model at training time,
    \item We demonstrate the competitiveness of \acosp with other pruning approaches by applying it to two widely used image segmentation model architectures trained on the \camvid and \city benchmark datasets,
    \item We show that the overwhelming majority of pruned filters is chosen at random, demonstrating the importance of \acosp's auto-compression and conclude with a study of the interaction of pruning and training.
\end{itemize}

The remainder of the paper is structured as follows: In \Cref{sec:related_work}, we give an overview of related pruning approaches. \Cref{sec:background} lays out the fundamentals of pruning and subset selection. \Cref{sec:parallel_subset_selection} introduces our approach and we demonstrate its application to semantic segmentation models. In \Cref{sec:experiments}, we show experiments and discuss results and analyses before we conclude our work in \Cref{sec:conclusions}.

\section{Related work}
\label{sec:related_work}

Parameter reduction techniques, such as pruning fully connected neural networks (FCN) in an unstructured way, date back to the early days of neural networks with seminal works of LeCun et al.~\yrcite{lecun_optimal_1990} and Hassibi \& Stork~\yrcite{hassibi_second_1992}. Since then, many advances have been achieved. Neill~\yrcite{neill_overview_2020} reviews various techniques. To give a few examples, Denil et al.~\yrcite{denil_predicting_2014} exploit that parameters in FCN and CNN exhibit structures that enable prediction of 95\% of the model's weights using only 5\% of the trained weights. In a different approach to pruning FCNs and CNNs, Srinivas \& Babu~\yrcite{srinivas_data-free_2015} exploit redundancies in the models by removing individual neurons rather than individual weights from the network. Lebedev \& Lempitsky~\yrcite{lebedev_fast_2015} build upon these works and aim at preserving the structural advantages of having few but dense layers. They propose to remove individual filters in a convolution based on their matrix multiplication form in conjunction with approaches developed in LeCun et al.~\yrcite{lecun_optimal_1990}. Li et al.~\yrcite{li_pruning_2017} simplify the objective function to prune convolutional filters based on their $\ell_1$-norm to induce thin networks. To regain lost accuracy, they introduce a second retraining step. Finally, a different route to compression of CNNs is taken by Liu et al.~\yrcite{baoyuan_liu_sparse_2015}, who reduce the parameter counts in a network by exploiting the redundancy in parameters by using a sparse decomposition of the convolutional kernels. This allows them to express the full network with a significantly reduced memory footprint.

The above methods rely on removing parameters post-training and regaining lost accuracy through few-step retraining. To the best of our knowledge, Liu et al.~\yrcite{liu_learning_2017} are the first to introduce a method that incorporates pruning into the training procedure. By using a filter-wise $\ell_1$-regularizer during training they identify insignificant filters. Bejnordi et al.~\yrcite{bejnordi_batch-shaping_2020} pick up the idea of dynamic pruning and learn a gating function that switches on and off the convolution operator on certain filters of the convolution dynamically based on the feature input to the convolution block. They demonstrate the feasibility on classification as well as segmentation tasks. A similar approach is followed by Kim et al.~\yrcite{kim_plug-_2019} who are learning a discrete rather than continuous gating function by introducing a relaxation of a non-differentiable discrete loss. Luo \& Wu~\yrcite{luo_autopruner_2019} predict the next layer's gating function from the previous layer's activations instead of the feature maps. Using a small CNN, they predict a binary variable that switches the computation on the next layers' convolution filters. Su et al.~\yrcite{su_data_2020} use the weights of a pre-trained model to learn a gating function of the next layer. This lets them also optimize for the computational budget of the model through an added loss term on the objective function. A commonality of all these approaches is their reliance on additional loss terms in the objective function or on discrete, non-smooth removals of whole filters. This can lead to sub-optimal solutions or shocks in the model's training dynamics.

Most work studying pruning for computer vision focuses on image classification \cite{Blalock2020} leaving the question of transferability of these results to pixel-wise classification, as done in semantic segmentation, largely open.  In contrast to the high information redundancy of classification-labels, semantic segmentation needs to preserve detailed pixel-level structures in the prediction map. When pruning indiscriminately, these structures can be lost, resulting in a lower of performance of the model, which might not be recoverable with re-training~\cite{Frankle2020The}. We are aware of three works that extend the investigations of pruning to semantic segmentation networks: He et al.~\yrcite{he_cap_nodate} introduce a context-aware guidance module that leverages local context-aware features to control the channel scaling factors in the batch-normalization layers. They demonstrate the approach's efficacy on a set of different model architectures trained on \city and \camvid. Schoonhoven et al.~\yrcite{lean_pruning} view CNNs as computational graphs and use a longest-path algorithm to find relevant chains of computation greedily. They train models on a synthetic dataset and \camvid. However, their reported results are not directly comparable to channel pruning, as they prune individual filter maps of a convolution rather than output channels.
Finally, You et al.~\yrcite{you_gate_2019}, introduce a channel-wise scaling factor whose importance is estimated through sensitivity analysis of the loss-function w.r.t., these parameters. They train these models on the \textsc{Pascal VOC2011} dataset \cite{pascal-voc-2011} by iterating between training the model for a fixed set of iterations and a sensitivity estimation followed by a sparsification phase.

\begin{figure}[t]
    \centering
    \includegraphics[width=0.55\columnwidth]{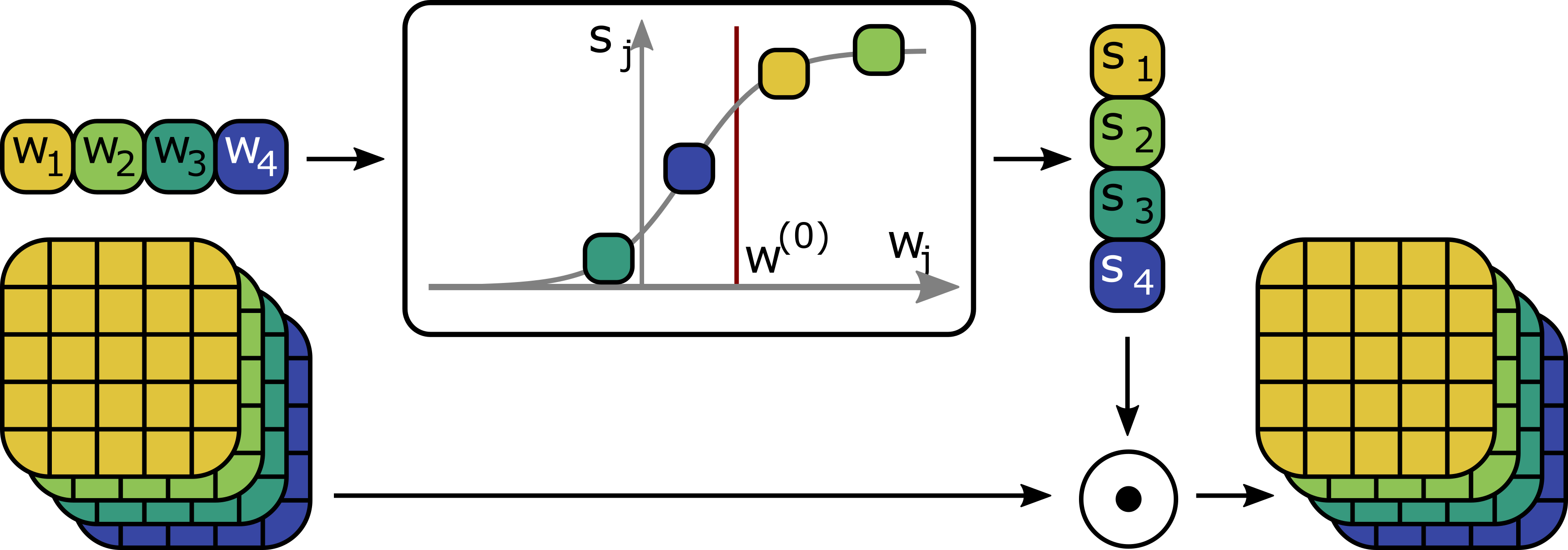}
    \caption{\acosp is applied to every convolution layer. It multiplies each output channel by a scaling factor $s_j$ parameterized by a logit $\omega_j$. An additional offset ensures that a predefined number is selected. Using temperature variable is gradually decreased annealing the \sigmoid to a step function until a binary channel selection mask remains.}
    \label{fig:schematic}
\end{figure}

\section{Auto-Compressing Subset Pruning}
\label{sec:background}

In contrast to the previous approaches, \acosp learns channel-importance weights during training through annealing, thus avoiding shocks in the system. The dependency on the annealing schedule is weak and we show that it can even be replaced with an arbitrary subset of filters to keep while slowly annealing all others weights to zero. Moreover, \acosp does not rely on additional loss functions that compete with the original objective and removes the complicated trade-off of weighting these losses against each other. As an additional benefit, \acosp also allows for fine-grained control of the final compression ratio. While we study \acosp in the context of segmentation, it should be noted that the method is task-agnostic and can readily be applied to other problems.

These results are achieved by using a modification of Differential Subset Pruning introduced in the context of pruning transformer heads in large NLP models~\cite{li_differentiable_2021}. We exchange the \softmax weighting with a simple \sigmoid and shifting operation to simulate an effective \topk operation and show how to apply this approach to CNN-based segmentation networks. This is in contrast to Xie et al.~\yrcite{xie_differentiable_2020} who are using an optimal-transport-based differentiable \topk relaxation at a much higher computational cost.

\subsection{Introduction}

We are focusing on four key elements that need to be considered when pruning: (i) the inductive bias of the final compressed model, (ii) the criteria to decide what should be pruned, (iii) the schedule when to prune the elements, and (iv) how much should be pruned. To briefly recap:

\textbf{Pruning structure.} To facilitate easy deployment on modern, specialized hardware, such as GPUs or TPUs, \acosp aims at preserving the model structure to create thin, but densely connected networks. The trade-off compared to unstructured pruning, is a higher drop in performance, which is typically acceptable given the hardware constraints and increased ease of deployment.

\textbf{Pruning criteria.} We avoid the introduction of additional loss functions based on heuristics, such as the $\ell_1$-weight norm, by choosing a learned pruning approach. \acosp learns a gate vector $\textbf{s}$ whose number of elements corresponds to the number of convolution output channels for each convolution in our segmentation network (see \Cref{fig:schematic}). If certain components of the gate are almost closed, subsequent layers depend on these features only weakly, if at all. By forcing the gate signal to be close to zero or one, we remove or retain the respective filters, effectively compressing the model into the filters with open gates.

\textbf{Pruning schedule.} To determine when to prune the chosen elements, we have several options: pruning before, after, or during the training and interleaving them in so-called tick-tock schedules~\cite{you_gate_2019}.

Removing weights in a structured fashion after training, introduces shocks into the system, due to the removal of important weights of the underlying, unstructured \textit{lottery winner}~\cite{frankle_lottery_2019}. This is associated with strong performance degradation, which can only partially be recovered through retraining. To avoid such shocks, \acosp compresses the model during training to transfer the information of the unstructured lottery winner into the remaining dense and structured filters. This is achieved by forcing the gating vectors to approach a binary value in $\{0, 1\}$ by using a \sigmoid-parameterization (see \Cref{fig:schematic}) and an effective temperature schedule.

\textbf{Compression ratio -- subset selection.} Typical online pruning techniques enforce compression ratios via additional loss functions. As a consequence, the final compression ratio is non-deterministic and hard to set at the outset of training. Posthoc methods, on the other hand, can easily be tailored to a target compression ratio, but suffer from the pruning shock as outlined above. \acosp addresses this by approaching structured pruning via a subset selection process. Given a finite set $\mathcal{E}= \{1,\dotsc, N\}\subset \mathbb{N}$ of elements and a pruning target fraction of $0<r<1$, the operation of removing $r\cdot N$ elements is equivalent to the selection of $K= \lfloor(1-r)\cdot N\rfloor$ remaining elements. Hence, pruning to the given fraction $r$ is equivalent to selecting a subset  $\mathcal{S}\subseteq \mathcal{E}$ of cardinality $K$ that maximizes the objective function. In the following, we will refer to this as a \topk operation. 

Mathematically, we express the selection of element $j\in \mathcal{S}$ through the scaling factors $s_j\in \{0,1\}$ and an additional constraint equation:
\begin{equation}
    j\in \mathcal{S} \Leftrightarrow s_j= 1,
    \text{ subject to }\label{eq:elem_sel} 
    \sum_j s_j= K.
\end{equation}
Solving this constraint optimization problem while maximizing the objective function allows us to compress the model by selecting $K$ individual elements under the observation of the total number of allowed selections. In our approach, this can be achieved by shifting the \sigmoid operation accordingly as indicated in \Cref{fig:schematic}.

\begin{algorithm}[t]
  \caption{\acosp}
  \label{alg:acosp}
\begin{algorithmic}
  \STATE {\bfseries Input:} data $x_i$, model $\mathcal{M}$, epoch $E$, temp. schedule $T$
  \STATE {\bfseries Output:} train \& pruned model $\mathcal{M}$
  \FOR{$e$ in $[1, \dots, E]$}
        \STATE $\tau \leftarrow T(e)$
        \FOR{\textsc{Conv}  $c$ in $\mathcal{M}$}
            \STATE $\omega_{c,j} \leftarrow $ \textsc{GetFilterWeight}($c$)
            \STATE $\omega^{(0)}_c \leftarrow  $ \textsc{Quantile}$_K$($\omega_{c,j}$)
            \STATE $s_{c,j} \leftarrow $ \sigmoid(($\omega_{c,j}- \omega^{(0)}_c$)  / $\tau$)
        \ENDFOR
        \STATE $\mathcal{M}$.update()
 \ENDFOR
\end{algorithmic}
\end{algorithm}

\begin{figure*}[t]
    \centering
    \includegraphics[width=0.98\linewidth]{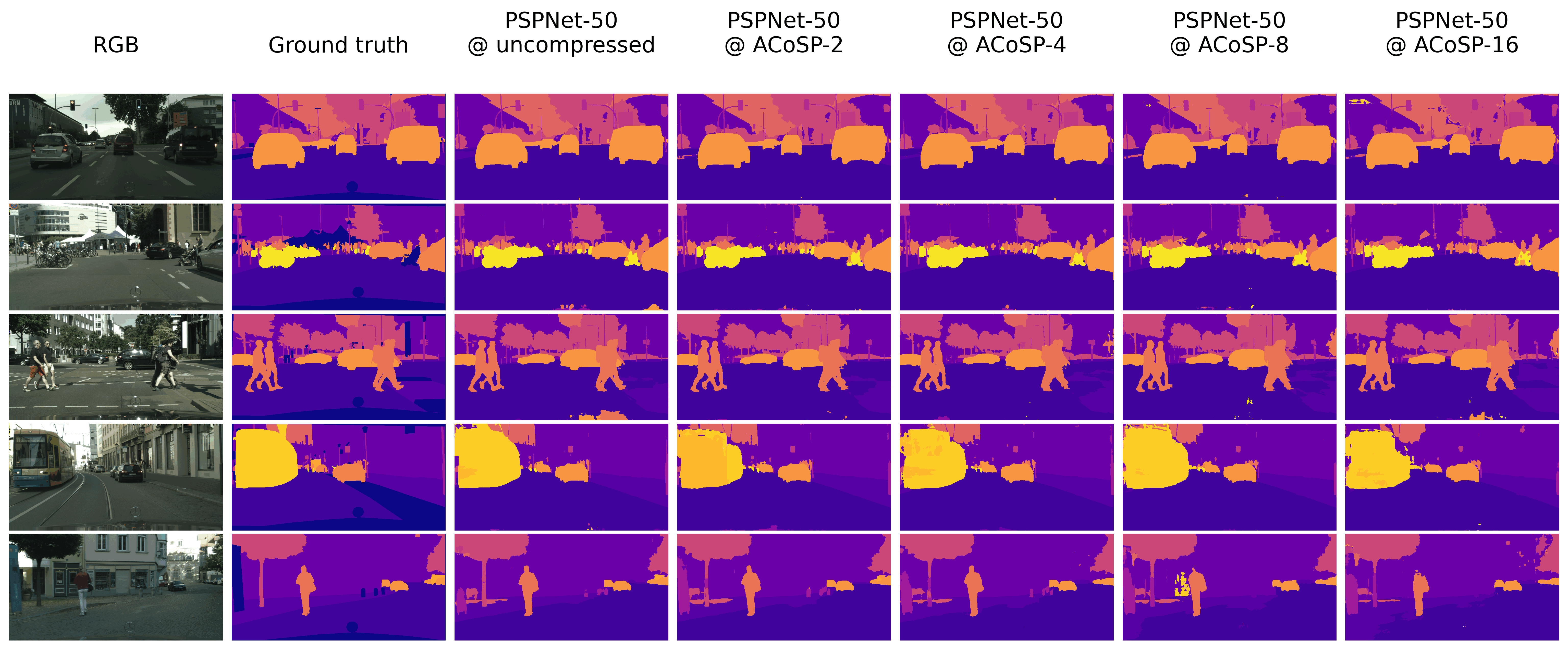}
    \caption{Impact of the \acosp strategy on the performance of a \pspnet-50, trained on \city, compared to the ground-truth at varying compression ratios. As can be expected, compression mostly impacts the resolution of fine details (clearly seen in the first row in the blue overhead bar and the vanishing of traffic poles in the second row). However, it is encouraging to see that the \acosp does not forget concepts. For instance, the person crossing the street is discovered at all compression ratios.}
    \label{fig:segmentation_degradation}
\end{figure*}

\subsection{Details of the algorithm}
\label{sec:parallel_subset_selection}
Leveraging subset selection to compress models efficiently is a general technique not limited to segmentation. We first describe it in an abstract setting and apply it to CNNs in a subsequent step.

\textbf{General framework of \sigmoid-based subset selection.} A first challenge is to find a differentiable relaxation of the \topk operator. While there are existing derivations of such operators \cite{li_differentiable_2021, xie_differentiable_2020, grover2019stochastic}, we have found them to be numerically unstable and not sufficiently efficient for our purposes, since pruning channels of CNNs involves a large number of elements to select from. We, therefore, introduce a fast, simple, stable, and differentiable approximation of the \topk operator that is applicable to a large set of elements.

We introduce the scaling factors $s_{c,j} \in [0,1]$, $j\in\{1, \dots, N_c\}$, $c\in\{1, \dots, C\}$, where $N_c$ is the number of elements in a layer $c$, e.g., channels in the case of convolution,  and $C$ is the total number of layers in the model. The $s_{c,j}$ are parameterized via learnable weights $\omega_{c,j}$ and constrain them to the unit interval by using the \sigmoid function (see \Cref{fig:schematic}). We introduce a temperature variable $\tau \in \mathbb{R}^+$ to achieve gradual annealing of the \sigmoid selection into a binary mask. In the limit of $\tau \longrightarrow 0$, the \sigmoid turns into the Heaviside-step function $\Theta(\omega_{c,j}) = \mathbf{1}_{\omega_{c,j}> 0}$. Here, $\mathbf{1}_{x}$ is the indicator function of condition $x$. Since \sigmoid is differentiable, gradients can be back-propagated through the gating operation, and the $\omega_{c,j}$ can be learned via gradient descent. The relaxed \topk operation is achieved by determining the offset value $\omega^{(0)}_{c}$, such that $\sum_{j}^N \mathbf{1}_{\omega_{c,j}>\omega^{(0)}_{c}} = K$. This is an extreme value problem, but it can easily be solved using a quantile function, or by selecting the $K$-th value of a reversely sorted array. The offset value allows us to shift the weights, and hence the scaling factors, in such a way that only the \topk elements fulfill 

\begin{align}
    \tilde{\omega}_{c,j} = \omega_{c,j}- \omega^{(0)}_{c}
    \begin{dcases}
        >0,              & \text{if } j\in\mathcal{S}\\
        <0,              & \text{otherwise}.
    \end{dcases}
\end{align}

Putting it together with the annealing schedule induced by the effective temperature, we can hence learn the channel selection mask via

\begin{equation}
      s_{c,j} = \sigmoid\left(\frac{\tilde{\omega}_{c,j}}{\tau} \right) \stackrel{\tau\rightarrow 0}{\longrightarrow}\begin{dcases}
        1,              & \text{if } j\in\mathcal{S}\\
        0,              & \text{otherwise}.
    \end{dcases}
\end{equation}

The full algorithm is summarized as pseudocode in Algorithm \ref{alg:acosp}. The key is to adjust the offset in each step to fix \topk elements that fulfill the Heaviside condition and slowly anneal the temperature to zero. The free choice of $K$ allows us to control the final compression ratio during training and, more importantly, does not require any hyper-parameters.

While the gating factors $s_{c,j}$ are differentiable with respect to the weights $\omega_{c,j}$, the operation is not fully differentiable, because $\omega^{(0)}_c$ depends on all $\omega_{c,j}$ in a non-differentiable way. However, we find that in practice, $\omega^{(0)}_c$ is not changing rapidly and can be regarded as almost constant.

\begin{table*}[t]
\fontsize{8}{10}\selectfont
\caption{Detailed results of our \acosp compressing strategy on the \city \cite{cordts_cityscapes_2016}, \voc \cite{pascal-voc-2012} and \ade \cite{Zhou2017ScenePT} datasets. Unless otherwise noted, all experiments use a PSPNet with a \resnetfive backbone pretrained on Imagenet \cite{deng2009imagenet} (for detailed results on \segnet we refer to the appendix). For \city, we report the mIoU on the test set if available and the validation set in parenthesis. The table clearly shows the strength of \acosp to keep a high performance despite a high compression ratio. $^\dagger$: Numbers are taken from \cite{he_cap_nodate}. $^*$: Numbers taken from Bejnordi et al.~\yrcite{bejnordi_batch-shaping_2020}; Numbers were calculated on the \textsc{Cityscapes} validation set.; While not technically compression the network, they do reduce the required MACs due to their gating process. Numbers are reflecting the performance with the gating mechanism and without the mechanism in parentheses. $^\#$: Numbers taken from You et al.~\yrcite{you_gate_2019}; the numbers are not directly comparable due to architectural differences (FCN-32 with \vgg backbone compared to a PSPNet50) as well as training being done on the \textit{extended} \textsc{Pascal VOC 2011} dataset \cite{pascal-voc-2011, Hariharan-2011}.
}
\centering
\begin{tabular}{|c|c|c|c|c|c|c|c|} 
\hline
  \textbf{Method} &
  \multicolumn{2}{c|}{\textsc{Cityscapes}} & 
  \multicolumn{2}{c|}{\textsc{Pascal VOC 2012} $^\#$} &
  \multicolumn{2}{c|}{\textsc{Ade20K}} \\
 \cline{2-7}
                         & mIoU(\%)     & Params[M]($\downarrow \%$) & mIoU(\%)      & Params[M]($\downarrow \%$) & mIoU(\%) & Params[M]($\downarrow \%$) \\
\hline
Unpruned (OURS)          & 77.37 (78.0) & 49.08                      & 72.71         & 49.08                   & 41.42    & 49.18                   \\
CCGN (from scratch) $^*$ & 71.9 (70.6)  & $(23.7\percdown)$          &               &                         &          &                         \\
CCGN (pretrained) $^*$   & 74.4 (73.9)  & $(23.5\percdown)$          &               &                         &          &                         \\
GBN $^\#$                &              &                            & 62.84 (62.86) & $36.5$ $(27\percdown)$     &          &                         \\
FPGM $^\dagger$          & 74.59        & $27.06$ $(47.40\percdown)$    &               &                         &          &                         \\
NS-20\% $^\dagger$       & 73.57        & $23.61$ $(54.11\percdown)$    &               &                         &          &                         \\
BN-Scale-20\% $^\dagger$ & 73.85        & $23.59$ $(54.15\percdown)$    &               &                         &          &                         \\
CAP-60\% $^\dagger$      & n/a (75.59)  & $27.31$ $(46.92\percdown)$    &               &                         &          &                         \\
CAP-70\% $^\dagger$      & n/a (73.94)  & $23.78$ $(53.78\percdown)$    &               &                         &          &                         \\
\acosp-2 Ours)         & 74.57 (74.1) & $24.57$ $(50.00\percdown)$    & 72.70         & $24.57$ $(49.94\percdown)$ & 38.97    & $24.62(49.94\percdown)$ \\
\acosp-4 (Ours)       & 71.05 (71.5) & $12.32$ $(74.90\percdown)$    & 65.84         & $12.32$ $(74.90\percdown)$ & 33.67    & $12.34(74.91\percdown)$ \\
\acosp-8 (Ours)      & 66.63 (66.1) & $6.19$ $(87.39\percdown)$     & 58.26         & $6.19$ $(87.39\percdown)$  & 28.04    & $6.20(87.40\percdown)$  \\
\acosp-16 (Ours)     & 59.50 (59.5) & $3.14$ $(93.60\percdown)$     & 48.05         & $3.14$ $(93.61\percdown)$  & 19.39    & $3.14(93.61\percdown)$  \\

\hline
\end{tabular}
\label{table:1}
\end{table*}

\textbf{Subset selection for CNNs.} The subset selection approach is general and can be applied to all problems involving a \topk operation. As a demonstration, we apply it to CNNs. As depicted in Fig. \ref{fig:schematic}, we introduce a channel-wise scaling factor, i.e., a learnable gating vector whose elements are the weights for scaling values for each channel. For simplicity of the implementation, we note that due to the linearity of the convolution operator, it is sufficient to multiply the activations of the convolution with the corresponding scaling factors rather than the weights of the convolution. It is important that no other operation (e.g., nonlinearity or normalization) happens between the convolution and the scaling to preserve the linearity of that operation.

We determine the desired compression by assigning each convolution in the network a number of remaining channels $K_c \in \mathbb{Z}^+$ and lower-bound it by a minimal number of channels $K_\text{min}$. We choose $K_\text{min}=8$ in our experiments. We then define a weight vector $w_c$ to represent the gate, containing the weights $\omega_{c,j}$. The convolution blocks of the models in our experiment follow the post-activation pattern \textsc{Conv} $\rightarrow$ \textsc{BatchNorm} $\rightarrow$ \textsc{Activation}. In theory, a batch normalization layer placed right after the modified scaling-convolution inverts the scaling effect during training, as pointed out by ~\cite{bejnordi_batch-shaping_2020}. However, we do not notice a reversion of the gradual application of the mask during training. We hypothesize this could be either due to the added $\epsilon$  when normalizing by the variance of each channel or due to the running average statistics of batch norm layers.

\textbf{Initialization and temperature annealing schedule.} During the calculation of the scaling factors, we normalize the weights $\omega_{c,j}$ of our gate vector $w_c$ to avoid saturation of the \sigmoid function and to keep gradients finite and non-zero. This also allows us to use a single temperature value for every scaling operator in the network, even if the gradients are significantly different in magnitude. We initialize all $\omega_{c,j}$ by drawing values from a uniform distribution instead of using a constant value.

Finally, in order to encourage the transfer of concepts and the compression of models, we anneal the temperature over time. In the spirit of Li et al.~\yrcite{li_differentiable_2021}, we choose an exponential decaying temperature schedule to keep a constant selection pressure during training. The exponential decay is fully determined by the decay constant, or the \textit{pruning duration} for simplicity, which constitutes a hyper-parameter in our setup. After the final temperature has been reached, the \sigmoid is close to a step function and the masking layer has turned into a binary selection. The gate weights receive no gradient updates anymore, and the pruning selection is final.

\section{Experiments}
\label{sec:experiments}

We evaluate the performance of \acosp on the semantic segmentation benchmarks \camvid~\cite{Brostow2009SemanticOC}, \city~\cite{cordts_cityscapes_2016}, \voc~\cite{pascal-voc-2012}, and \ade~\cite{Zhou2017ScenePT}. For our experiments, we use the \segnet architecture with a \vgg backbone and a \pspnet with a \resnetfive backbone. Both backbones are pretrained on \imagenet~\cite{deng2009imagenet}.

To be able to evaluate \acosp, we choose He et al.~\yrcite{he_cap_nodate}, You et al.~\yrcite{you_gate_2019}, and Bejnordi et al.~\yrcite{bejnordi_batch-shaping_2020} as baselines, which motivates our choice of datasets and architectures. Most notably, this excludes state-of-the-art semantic segmentation architectures, and we leave the study of these architectures to future work. We note that we do not have access to either implementation or pretrained checkpoints of the references we compare ourselves to. Hence, we tried to reproduce these uncompressed baselines to the best of our knowledge. 

We train each model on the respective datasets using the following procedure (we refer the reader to \Cref{sec:implementation_details} for full details): We implement all our experiments in \textsc{PyTorch}~\cite{PyTorch_NEURIPS2019}. To train the models, we use the standard SGD optimizer with an initial learning rate of $0.01$, a momentum of $0.9$, and weight decay of $1e-4$.
For \pspnet, we base our implementation on \cite{semseg2019}. To train the segmentation head, we multiply its learning rate by $10$. The input data is normalized using the mean and standard deviation of \imagenet, according to standard practice. Additional random scaling, rotations, flipping, and Gaussian blurring transformations are used to augment inputs during training. Depending on the dataset, either the full input images are passed to the model or random crops of a fixed size.

We use \acosp to prune the filters of all convolution and deconvolution layers of the architectures, except for the final output layer to retain expressivity. A global compression ratio is used to compute the retained number of channels for each convolution. However, independent of the compression ratio, at least 8 channels are being kept. We found this minimal number of channels to be beneficial during training and of negligible impact on parameters and floating point operations (FLOPs).

We run multiple ablation studies varying compression ratios and pruning durations, and additionally analyze the learned filter selection. To evaluate our trained models, we calculate the mean Intersection-over-Union (mIoU). Here we follow the evaluation pattern of Zhao et al.~\yrcite{zhao_pyramid_2017} and use a sliding window over the full input image, followed by averaging the class probabilities and then taking the \textsc{argmax} to predict the pixel class. Finally, we analyze the influence of the pruning duration on the subset selection and the performance of the final model.

\textbf{Qualitative assessment.} \Cref{fig:segmentation_degradation} shows a subset of segmentation mask from a \pspnet trained on \city at varying degrees of compression (see \Cref{app:segmentation_masks} for masks of other models and datasets). The samples have been chosen explicitly to highlight the impact of \acosp on a variety of classes as well as small and large image features. \acosp is able to preserve the major class information even with a compression ratio of $16$ -- a reduction of more than one order of magnitude in numbers of parameters -- with degradations mostly limited to fine features such as sign posts and lamps. Moreover, it is important to note that \acosp does not result in forgetting concepts but rather only a loss in boundary precision. Since we train the models using cross-entropy losses there is reason to believe that some of this loss can be recovered by using a more tailored loss function~\cite{Jadon2020ASO, Reinke2021CommonLO} such as the \textsc{DICE}-loss of S{\o}renson~\yrcite{sorenson1948method}.

\textbf{Compression ratio.} A major benefit of \acosp is its ability to specify the compression ratio \textit{a-priori}, in contrast to methods that require implicit methods, such as additional loss functions.
We choose the compression ratios of our experiments $R=\left\{2^{k}\;\big\vert k=0,\dotsc,4 \right\}=\left\{1,2,4,8,16\right\}$ in a way that halves the remaining filters for every experiment in the series.

As can be seen in \Cref{fig:pruning_ratios}, we compare favorably to the baselines on the \city dataset for small compression ratios and outperform them significantly at high compression ratios for both architectures, \segnet and \pspnet. We report more exhaustive numbers in \Cref{table:1}, to show that this behavior is consistent across datasets. We report similar results on \segnet and \city in the appendix. The breadth of these results indicates that the approach is robust across model architectures and datasets and thus can be anticipated to generalize.

\begin{figure}[t]
    \centering
    \includegraphics[width=0.55\linewidth]{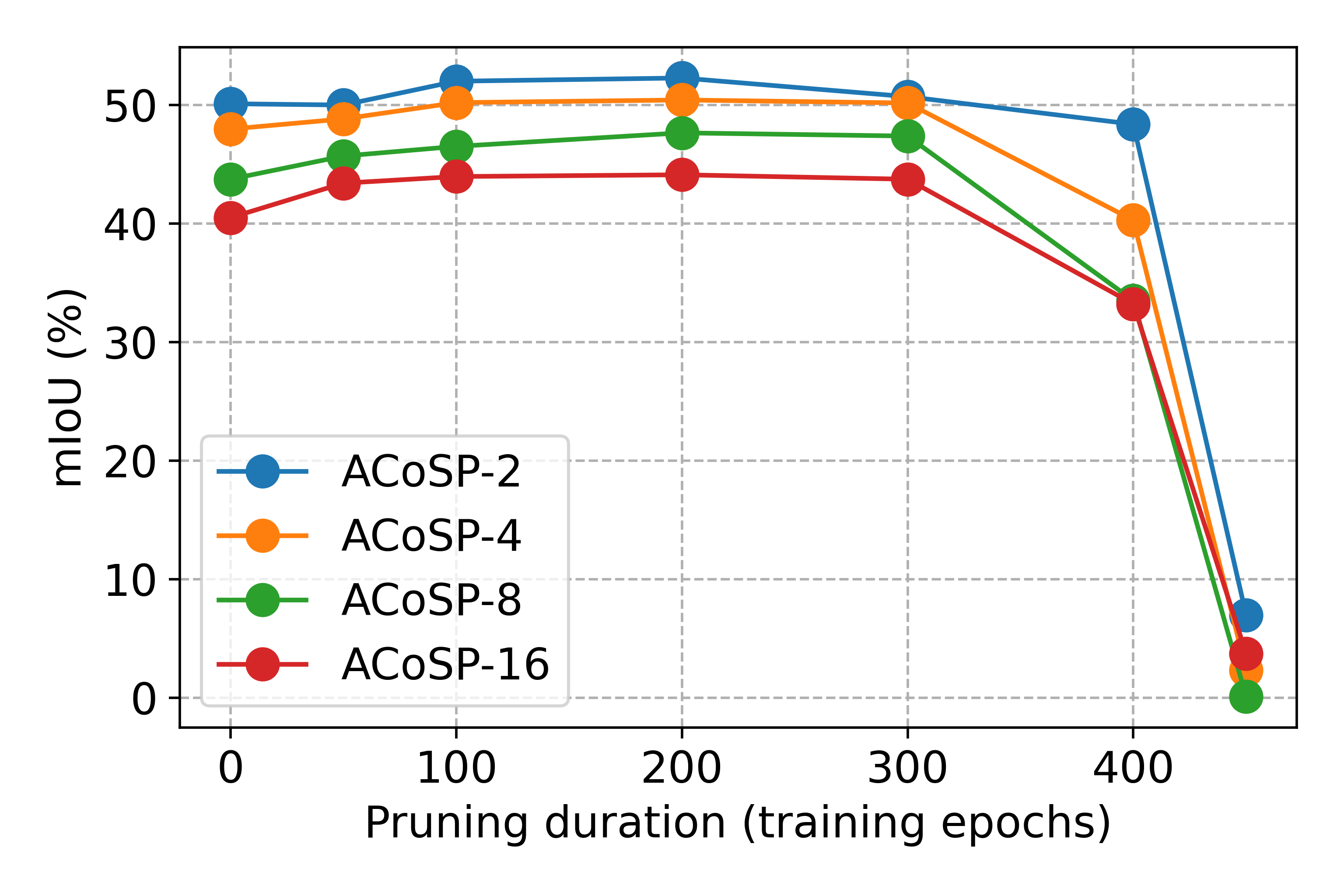}
    \caption{mIoU as a function of pruning durations. The results are obtained using a \segnet architecture trained on \camvid. The different curves show similar behavior for all tested compression ratios. Moreover, the performance of the final model is fairly robust w.r.t. the exact choice of the pruning duration. The performance drop due to the pruning shock of \textit{post-hoc} pruning is clearly visible, and a minor drop due to a-priori training as well. Latter is likely due to the limited available capacity of the model.}%
    \label{fig:pruning_durations}%
\end{figure}

\textbf{Pruning duration.} Since \acosp introduces the pruning duration as a new hyper-parameter via the temperature annealing schedule, we investigate the dependence of the results on the choice of this parameter. We fix the initial and final temperature at $1$ and $0.001$, respectively, and vary the annealing duration of the exponential schedule according to $D=\{0 ,50,100,200,300,400,450\}$. Note that $D=0$ corresponds to a-priori pruning, i.e., pruning at the start of the training, while $D=450$ corresponds to pruning after training, i.e., post-hoc compression. Intuitively the pruning duration controls the time the model has to compress relevant concepts into layers that are not pruned in the end.

Due to computational budget reasons, we perform this ablation by training a \segnet on the \camvid dataset. The results are shown in Figure \ref{fig:pruning_durations} and indicate that the approach is fairly robust with respect to the exact choice of the pruning duration. As expected, pruning at initialization has slightly inferior performance due to the restricted capacity of the model. It can be expected that a more complicated and/or larger dataset will show a significantly more pronounced effect in the a-priori pruning case due to under-parametrization. Similarly, pruning at the end of training introduces a big shock to the system, resulting in a significant drop in performance. Otherwise, \acosp is not sensitive with respect to the choice of the pruning duration. This behavior generally aligns with the intuition that the model needs the additional capacity to train efficiently in the beginning. Later in the training, the concepts can be gradually compressed into the remaining layers so that pruning irrelevant layers at the end of training does not introduce a shock.

\begin{figure}[t]
    \centering
    \includegraphics[width=0.55\linewidth]{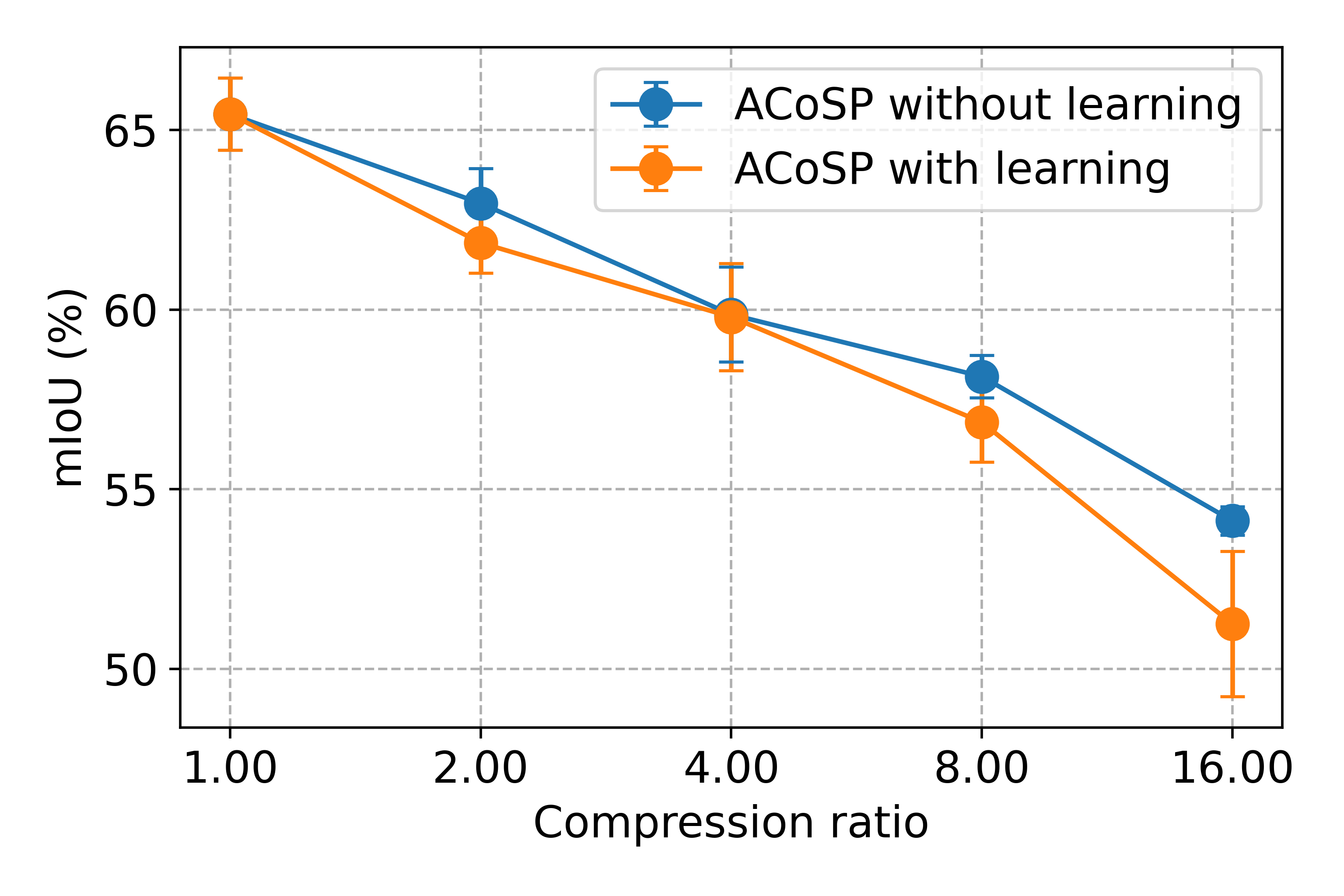}
    \caption{mIoU of \segnet trained on the \camvid training set comparing results with and without a learned selection. The models are evaluated on the \camvid test set. The experiments show little difference, indicating that the exact mechanism of how to choose the filters is not crucial, but rather that the initial capacity is necessary in addition to the compression pressure induced by the annealing schedule.}%
    \label{fig:pruning_annealing}%
\end{figure}

\textbf{\acosp without learning.} To estimate the importance of learning the gating weights, we run a control experiment where we initialize the weights as described before but simply do not update them during training. We keep them as fixed, non-trainable parameters. However, we do keep the temperature annealing schedule, and hence, we already know the exact filters that will be retained or removed at the end of the training. It is important to note that this is not equivalent to pruning at the beginning of the training, as the model has the capacity of the full model at its disposal at the beginning of training. The difference is the ability of the model to compress the learned concept into a pre-determined set of filters before the capacity is reduced at the end of training.

Again, due to computational constraints, we run this experiment using a \segnet model trained on \camvid. The results are shown in Figure \ref{fig:pruning_annealing} and show little difference between learning the filter selection versus choosing filters at random at initialization. While these results are not necessarily transferable to other datasets or model architectures, they still indicate that the exact gating mechanism is not crucial but rather that the models need capacity in the beginning of training in order to identify concepts quickly and then compress them.

\textbf{Generalization of the approach.} The mechanism behind \acosp is generally applicable to all CNN-based architectures and with more modifications also to other model architectures. Hence it can also be used for other tasks and is not specific to semantic image segmentation. To demonstrate the transferability, we also evaluate \acosp on the \cifar image classification task using \resneteight resulting in a drop from $89.71\%$ to $76.85\%$  top-1 accuracy when reaching the maximum compression level. As the results are qualitatively similar to results obtained for the semantic segmentation task, we choose to omit the discussion from the main body and refer the reader to the \Cref{sec:compression_ratios_extended} for more details.

\section{Summary and Outlook}
\label{sec:conclusions}

In this work, we focus on compressing semantic segmentation models to produce densely connected but thin architectures out of large state-of-the-art model architectures. The resulting models can easily be deployed in environments constrained by available compute and memory resources.

We introduce \acosp, a new online compression method that selects convolutional channels during training using a learnable gate vector parameterized by a shifted-\sigmoid layer in conjunction with a temperature annealing schedule. Major advantages of this method are (i) its ability to specify the compression ratio a-priori and (ii) its independence of additional loss objectives that need tuning while at the same time (iii) being easy to implement. This comes at the expense of an additional hyper-parameter. However, we demonstrate that the method is robust with respect to the exact choice of this parameter. We show that our approach is robust with respect to model architectures and dataset choices and is competitive with existing segmentation baselines in the low-compression regime. Moreover, \acosp outperforms previous approaches significantly in the high-compression regime on most datasets, with decent results even when discarding more than 93\% of the model. We show that the success of \acosp is not based on its exact selection mechanism but rather on the availability of large model capacity at the beginning of training working together with a compression pressure induced by the temperature annealing schedule driving the selection process.

While we studied this method in connection with semantic segmentation using CNNs, the underlying approach is generic. It can readily be applied to other data modalities and tasks and model architectures as demonstrated by a proof of concept using an image classification task. We leave further explorations as well as the application of \acosp to state-of-the-art architectures, different tasks and data modalities and the study of more tailored loss functions to future work.

Our work contributes to the field of online model compression methods and rephrases the process in terms of self-compression via concept transfer. It enables us to leverage the advantage of large, over-parameterized state-of-the-art models while at the same time being able to tailor the resulting model to a given resource-constrained environment. We hope this work inspires more research investigating the transfer of relevant unstructured model weights into unpruned subsets of a more structured model.

\section{Author contributions, Acknowledgements and Disclosure of Funding}
The authors would like to thank Thomas Wollmann, Stefan Dietzel, and Alexandra Lindt for valuable discussions and proofreading the manuscript. We also thank NVIDIA for generously providing us with compute resources as part of the partnership with the AI Campus, Berlin, as well as Altair for their IT support.

K.D. contributed to the design and implementation of the research, performed the experiments and analysis, and wrote the paper. J.S.O. contributed to the design of the research, wrote the paper, and supervised the work. All authors discussed the results and contributed to the final manuscript.

This work has been funded by EU ECSEL Project SECREDAS Cyber Security for Cross Domain Reliable Dependable Automated Systems (Grant Number: 783119). The authors would like to thank the consortium for the successful cooperation.

\bibliography{dsp}
\bibliographystyle{bib}
\newpage
\appendix
\onecolumn

% \begin{appendices}
\section{Implementation Details}
\label{sec:implementation_details}

In this section, we want to carefully explain our experimentation setup and increase reproducibility.
While the code is available at \url{\github}, a clear description of our usage and exact configurations is necessary to built upon our results.

\subsection{General settings.}

Due to the dataset and model-specific hardware requirements and compute constraints, we use three different hardware setups for our experiments.
\begin{itemize}
    \item A single NVIDIA T4 is used for all training on \camvid and \cifar datasets,
    \item otherwise 4 NVIDIA A100s are used in parallel to train models without \acosp,
    \item a single NVIDIA A100 is used per training with \acosp.
\end{itemize}

Experiments are conducted with either a \segnet or a \pspnet for segmentation tasks, while a \resneteight is used for classification.
The \segnet uses a \vgg backbone while the \pspnet uses a \resnetfive backbone.
To train the \pspnet segmentation head, we multiply its learning rate by $10$. 
Both are pretrained on \imagenet.

According to standard practice, the input data is z-normalized using the mean and standard deviation of \imagenet. Additional random scaling, rotations, flipping, and Gaussian blurring transformations are used to augment inputs during training.

Depending on the model architecture, either the full (augmented) input images are passed to the model or random crops of a fixed size.
Elaborate details on hyper-parameters can be found in Table \ref{table:hyperparams}.

\begin{table}[H]
\fontsize{8}{10}\selectfont
\centering
\caption{Detailed hyper-parameter overview. $^*$: \sgd refers to a stochastic gradient descent optimizer with a momentum of $0.9$ and weight decay of $1e-4$. While for \adam a weight decay of $5e-4$ is used. $^\dagger$: Either a cosine annealing schedule or a polynomial annealing schedule with an exponent of $0.9$ is used.}
\begin{tabular}{|c|c|c|c|c|c|c|} 
\hline
  \textbf{Hyper-Parameter} &
  \camvid + &  \city + &  \city + &  \voc + &  \ade + & \cifar + \\
 \textbf{Configurations} & \segnet   &  \segnet &  \pspnet &  \pspnet &  \pspnet & \resneteight \\
\hline
Training Epochs & $450$ & $200$ & $200$ & $50$ & $100$ & $50$    \\
Optimizer$^*$ & \sgd & \sgd & \sgd & \sgd & \sgd & \adam \\
Batch Size & $8$ & $16$ & $16$ & $16$ & $16$ & $512$ \\
Learning Rate Schedule$^\dagger$ & Cosine & Poly & Poly & Poly & Poly & Cosine \\
Initial Backbone Learning Rate & $0.01$ & $0.01$ & $0.01$ & $0.01$ & $0.01$ & $0.01$ \\
Initial Head Learning Rate & $0.01$ & $0.01$ & $0.1$ & $0.1$ & $0.1$ &  \\
\hline
Training Input Size & $360\times 480$ &  $512\times 1024$ &  $713\times 713$  & $473 \times 473$ & $473 \times 473$ & $32\times 32$\\
Evaluation Type & Full Image & Full Image & Sliding Window & Sliding Window & Sliding Window & Full Image \\
Evaluation Input Size & $360\times 480$ &  $1024\times 2048$ &  $713\times 713$ & $473 \times 473$ & $473 \times 473$  &  $32\times 32$\\ 
\hline

\end{tabular}

\label{table:hyperparams}
\end{table}

\section{Detailed results for \segnet.}

In addition to the results shown above, we provide detailed results for experiments using \segnet models in Table \ref{table:segnet}.
The models are trained on \camvid and \city.
While the results of \acosp are qualitatively similar for the \segnet architecture as for \pspnet, a significant performance gap exists on \camvid compared to the other approaches.
On \city, \acosp performs better than all available competitors again.
As \camvid is a significantly smaller dataset and arguably less complex than the other tested datasets, a possible reason is that other approaches are better at benefitting from extreme over-parametrization.

\begin{table}[H]
\centering
\tiny
\caption{
Detailed results of our \acosp compression strategy on the \camvid \cite{Brostow2009SemanticOC} and \city \cite{cordts_cityscapes_2016} datasets. All experiments use a \segnet with a \vgg backbone pretrained on \imagenet. For \city, we report the mIoU on the test set if available and the validation set in parenthesis.
($^\dagger$: numbers taken from \cite{he_cap_nodate})}
\begin{tabular}{|c|c|c|c|c|} 
\hline

\textbf{Method}          & \multicolumn{2}{c|}{\camvid} & \multicolumn{2}{c|}{\city}                 \\ \cline{2-5}
                         & mIoU(\%)                     & Params[M]($\downarrow \%$) & mIoU(\%)         & Params[M]($\downarrow \%$)  \\
\hline
Unpruned                 & $55.49\pm0.85$               & $29.45$                    & 66.61 (65.12)    & $29.45$                  \\
FPGM $^\dagger$          & $52.54$                      & $15.63 (46.92\percdown)$   & 51.60            & $15.63 (46.92\percdown)$ \\
NS-20\% $^\dagger$       & $54.78$                      & $12.25 (58.40\percdown)$   & 56.85            & $11.85 (59.76\percdown)$ \\
BN-Scale-20\% $^\dagger$ & $55.73$                      & $12.50 (57.39\percdown)$   & 59.95            & $11.92 (59.52\percdown)$ \\
CAP-20\% $^\dagger$      & $57.12$                      & $11.47 (61.05\percdown)$   & 61.16            & $10.76 (63.46\percdown)$ \\
CAP-30\% $^\dagger$      & $56.37$                      & $6.03 (79.52\percdown)$    &                  &                          \\
\acosp-2(Ours)        & $51.85\pm0.83$               & $14.74 (49.9\percdown)$    & 64.23 (64.62)    & $14.73(49.97\percdown)$  \\
\acosp-4 (Ours)       & $50.10\pm1.11$               & $7.38 (74.9\percdown)$     & 61.13 (60.77)    & $7.38(74.94\percdown)$   \\
\acosp-8 (Ours)      & $47.25\pm1.18$               & $3.71 (87.4\percdown)$     & 54.50 (54.34)  & $3.70(87.43\percdown)$   \\
\acosp-16 (Ours)     & $42.27\pm2.00$               & $1.88 (93.6\percdown)$     & 44.53 (44.12)  & $1.87(93.64\percdown)$   \\
\hline
\end{tabular}

\label{table:segnet}
\end{table}

\section{Compression Ratios}
\label{sec:compression_ratios_extended}

\begin{figure}[H]
\centering

\begin{subfigure}{.4\textwidth}
    \includegraphics[width=\linewidth]{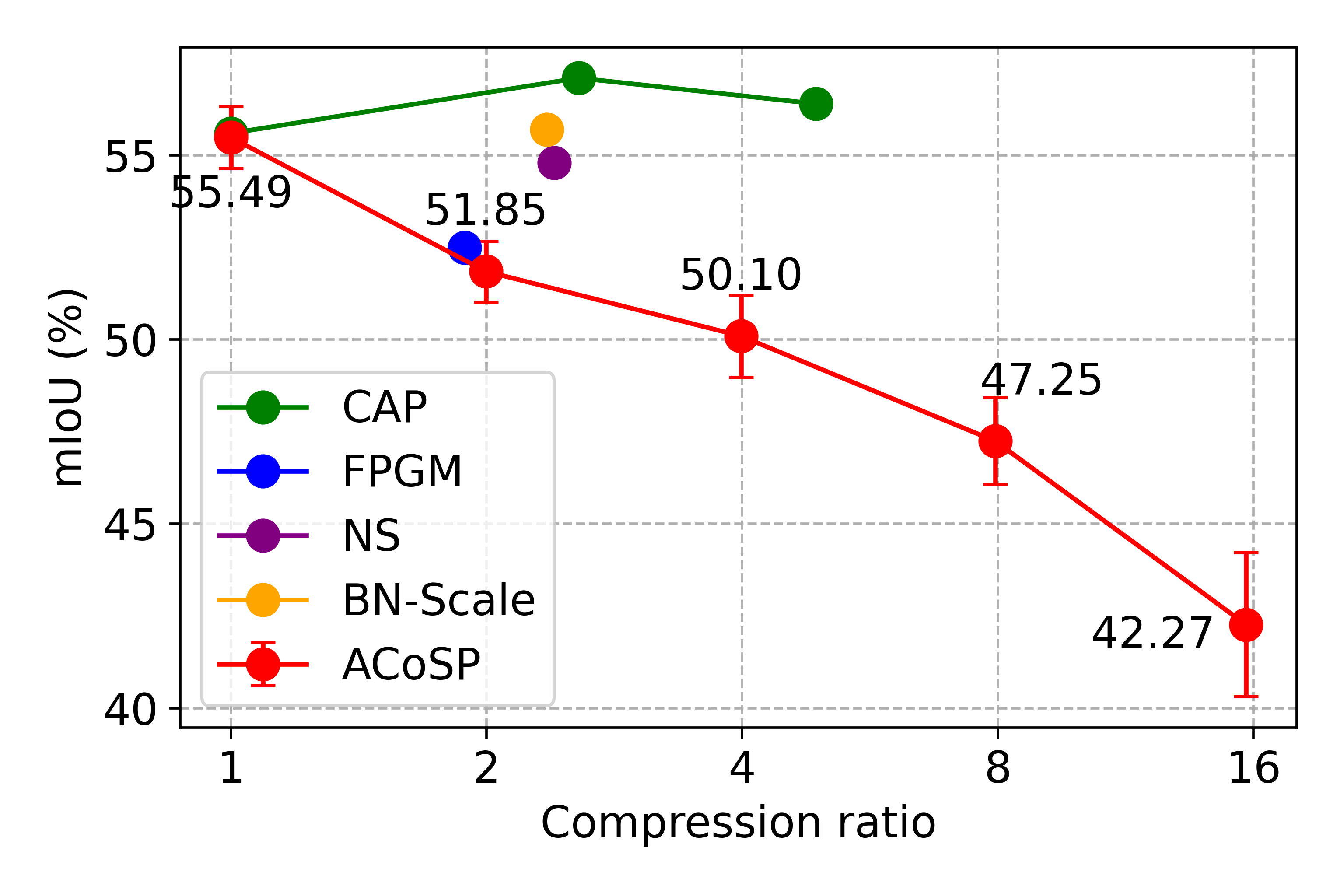}
    \caption{Mean Intersection-over-Union (mIoU) as a function of the compression ratio for \acosp by training \segnet models on the \camvid dataset.
    }
\end{subfigure}

\begin{subfigure}{.4\textwidth}
\includegraphics[width=\linewidth]{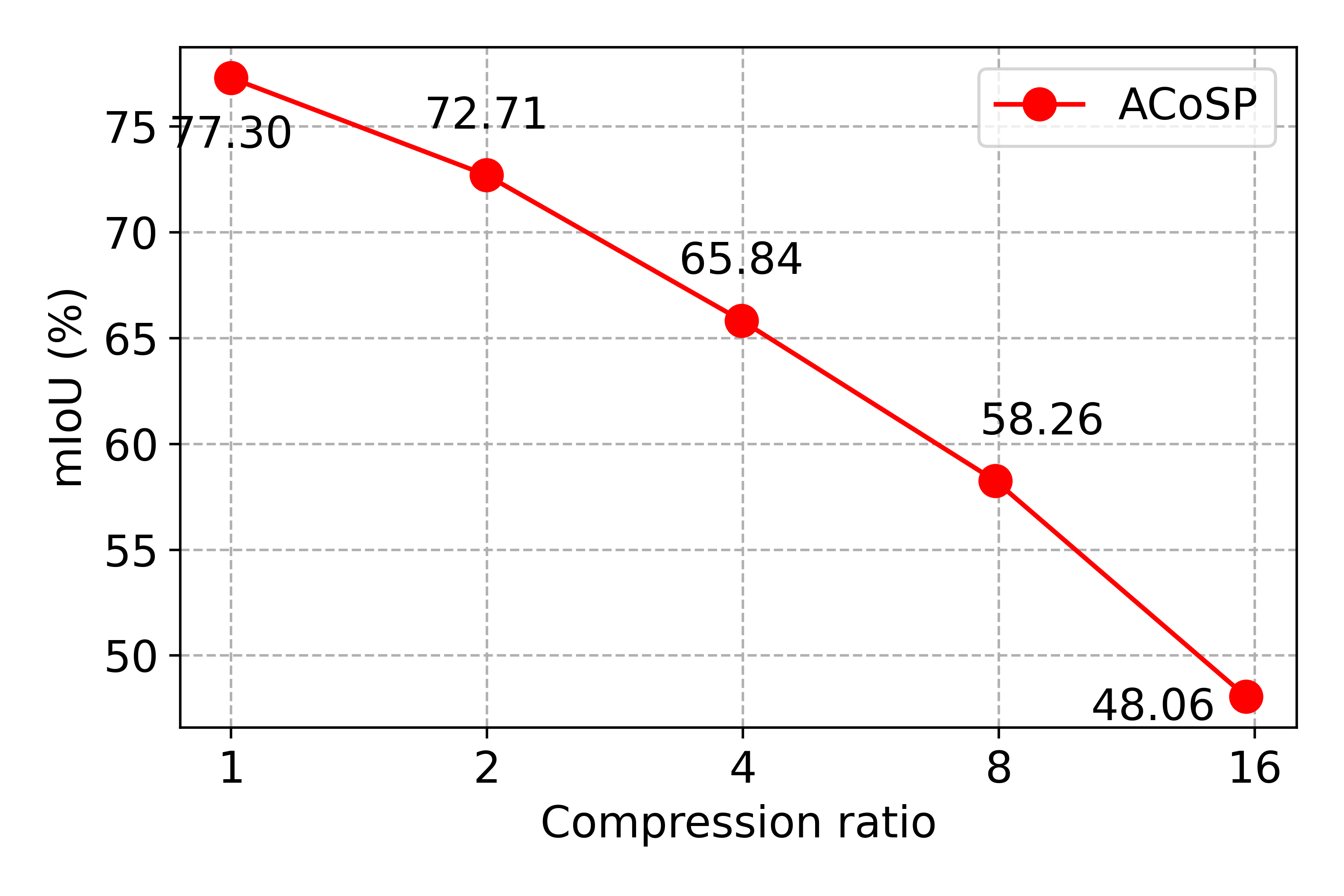}
    \caption{Mean Intersection-over-Union (mIoU) as a function of the compression ratio for \acosp by training \pspnet models on the \voc dataset.}
\end{subfigure}
\begin{subfigure}{.4\textwidth}
\includegraphics[width=\linewidth]{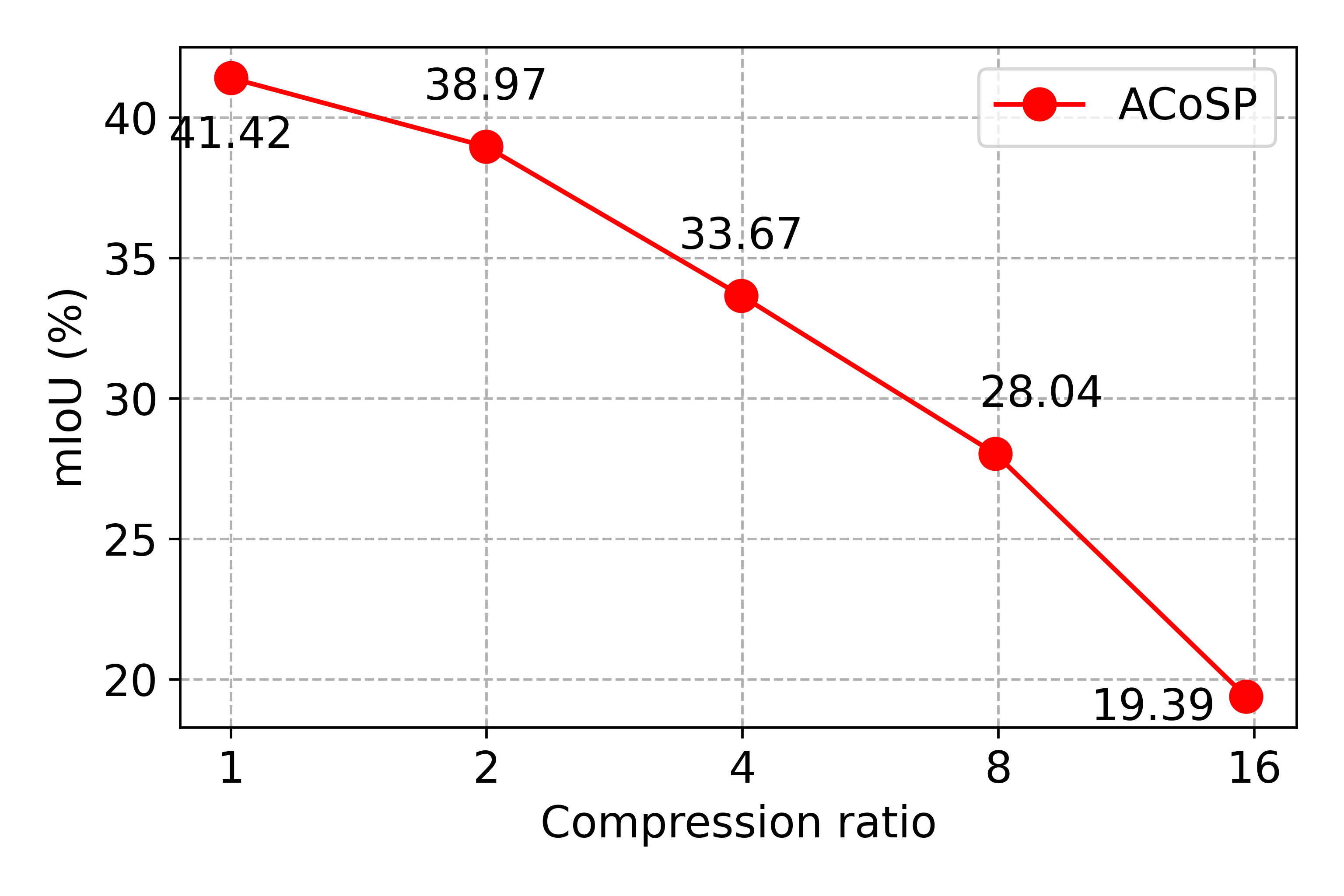}
    \caption{Mean Intersection-over-Union (mIoU) as a function of the compression ratio for \acosp by training \pspnet models on the \ade dataset.}
\end{subfigure}
\begin{subfigure}{.4\textwidth}
\includegraphics[width=\linewidth]{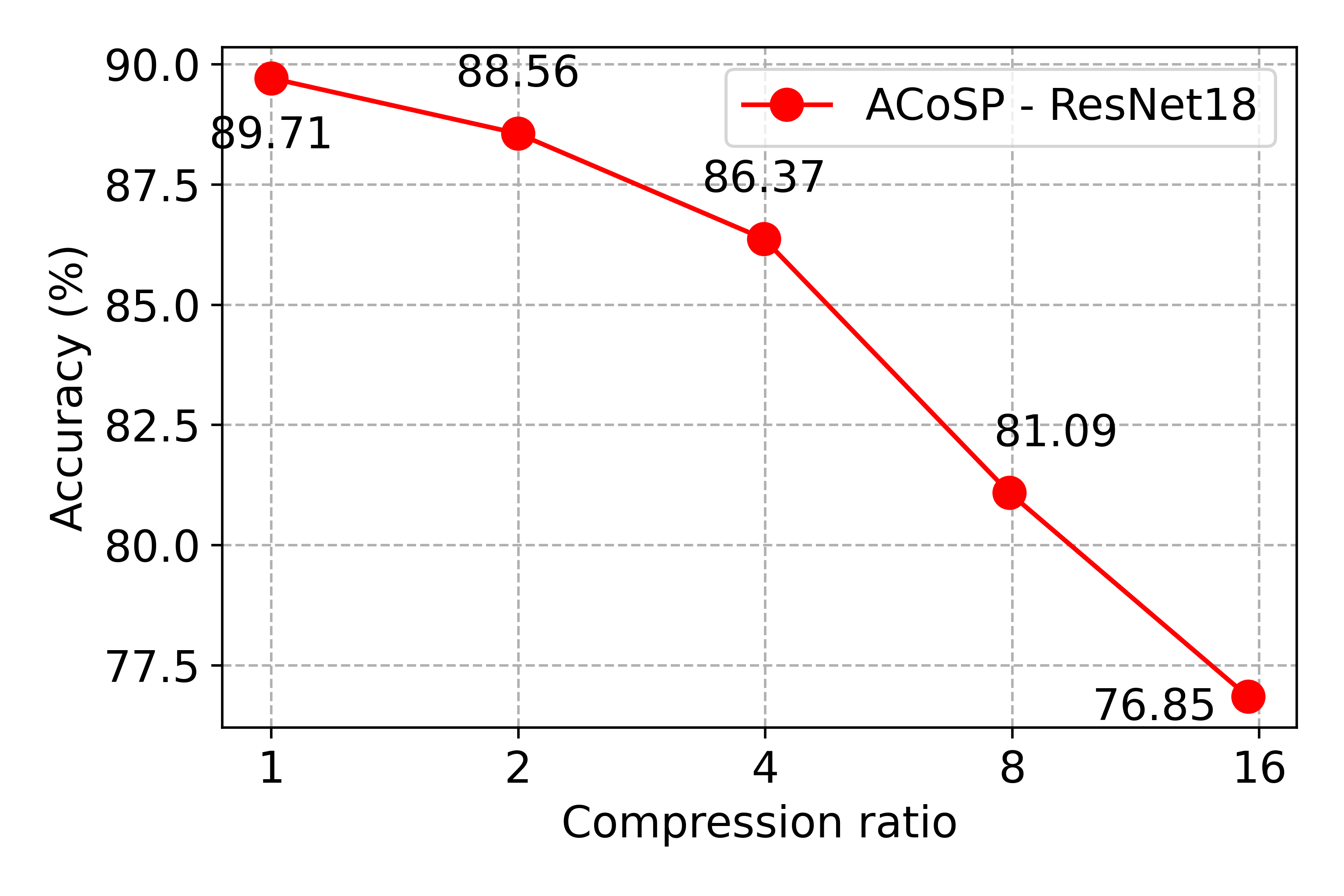}
    \caption{Top-1-Accuracy as a function of the compression ratio for \acosp by training \resneteight models on the \cifar dataset.}
\end{subfigure}
\caption{Final performance with respect to compression ratio of models trained with \acosp on multiple different datasets, architectures, and even computer vision tasks.}
\label{fig:pruning_ratios_missing}
\end{figure}

In addition to the previous plot of the compression ratio ablation study on \city, we want to show curves for the equivalent experiments with other model and dataset combinations.
\Cref{fig:pruning_ratios_missing} displays the qualitatively predictable performance decay of models trained with \acosp.

\section{Segmentation Masks}
\label{app:segmentation_masks}

\begin{figure}[H]
    \centering
    \includegraphics[width=0.98\linewidth]{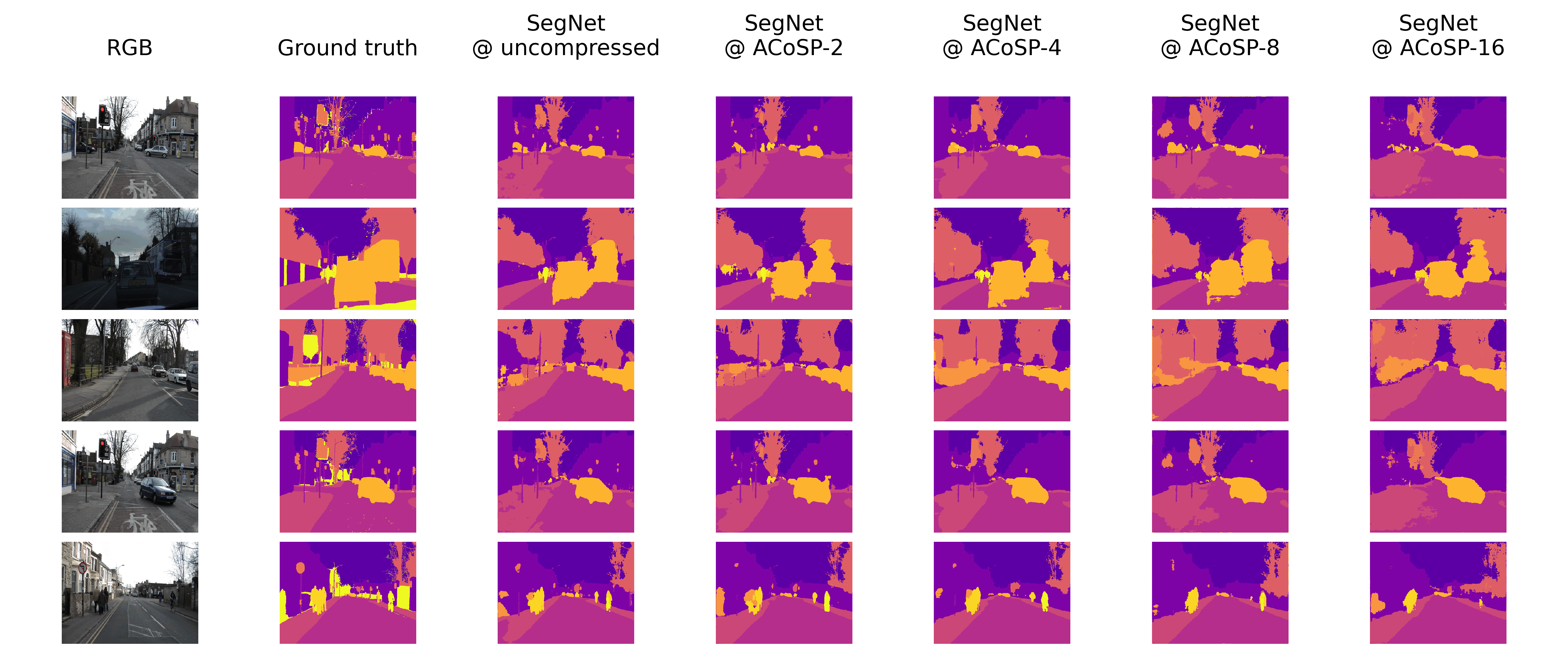}
    \caption{Impact of the \acosp strategy on the performance of a \segnet, trained on \camvid, compared to the ground-truth at varying compression ratios.
    }
    \label{fig:segmentation_degradation_camvid}
\end{figure}

\begin{figure}[H]
    \centering
    \includegraphics[width=0.98\linewidth]{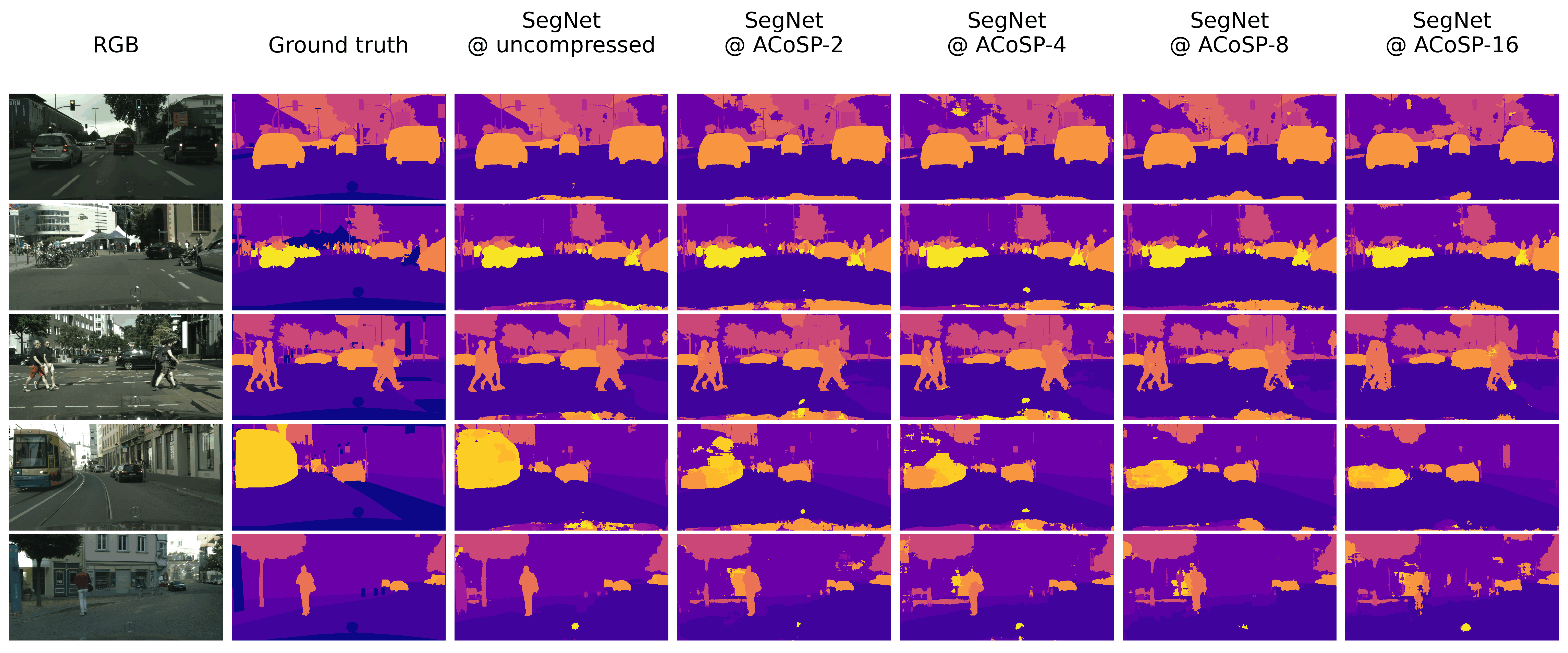}
    \caption{Impact of the \acosp strategy on the performance of a \segnet, trained on \city, compared to the ground-truth at varying compression ratios.}
    \label{fig:segmentation_degradation_city_segnet}
\end{figure}

\begin{figure}[H]
    \centering
    \includegraphics[width=0.98\linewidth]{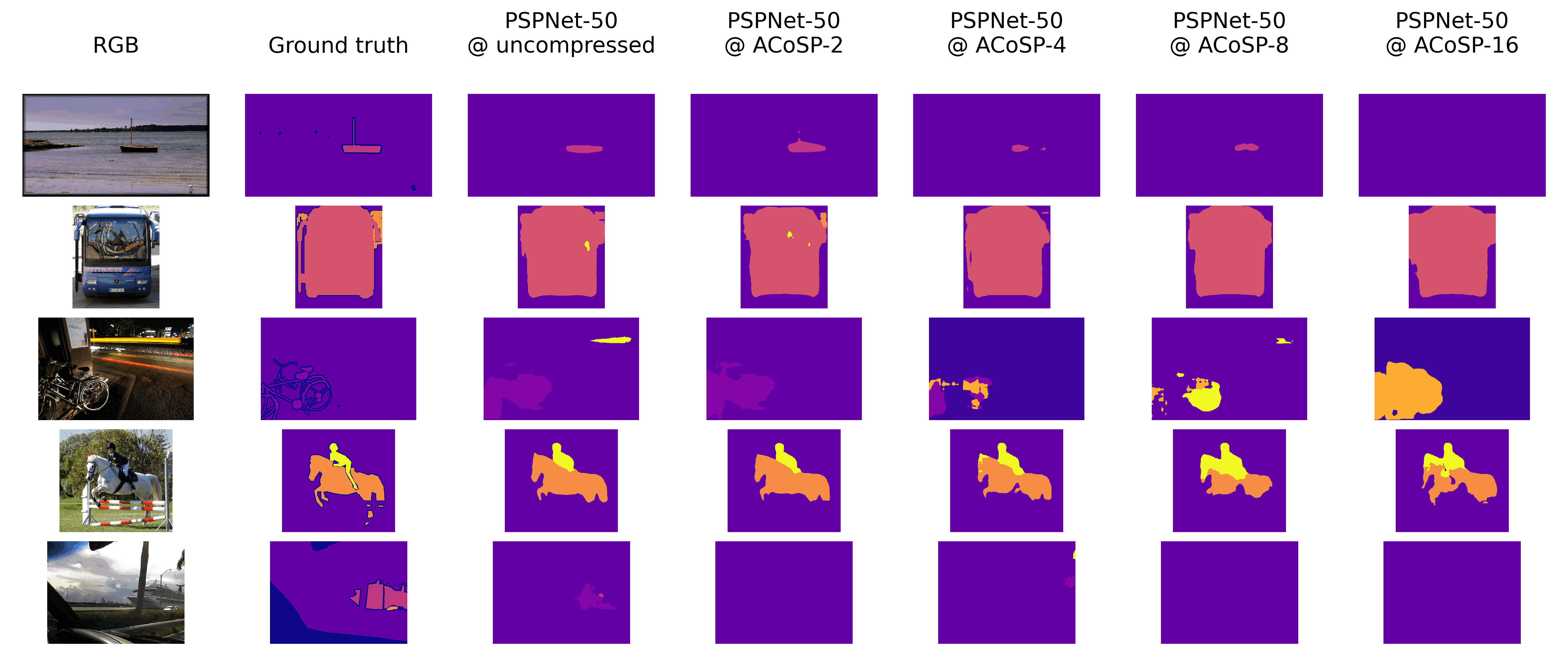}
    \caption{Impact of the \acosp strategy on the performance of a PSPNet-50, trained on Pascal VOC2012, compared to the ground-truth at varying compression ratios.}
    \label{fig:segmentation_degradation_voc}
\end{figure}

\begin{figure}[H]
    \centering
    \includegraphics[width=0.98\linewidth]{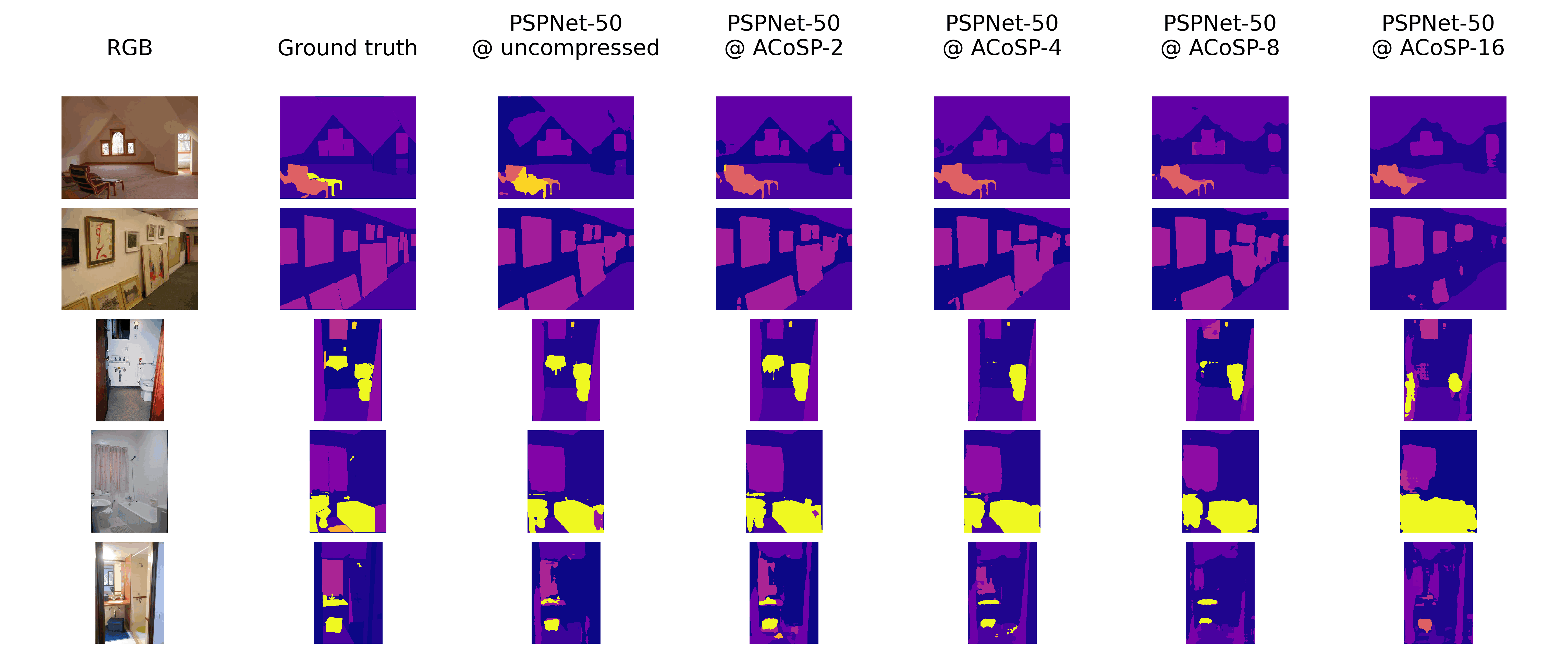}
    \caption{Impact of the \acosp strategy on the performance of a \pspnet50, trained on Ade20k, compared to the ground-truth at varying compression ratios.}
    \label{fig:segmentation_degradation_ade20k}
\end{figure}

Segmentation masks are an effective way of showing how the performance degradation of pruned models is distributed.
The segmentation masks in \Cref{fig:segmentation_degradation_camvid,fig:segmentation_degradation_city_segnet,fig:segmentation_degradation_voc,fig:segmentation_degradation_ade20k} provide qualitative insights into the degradation of our compressed models.

Throughout the experiments, degradation predominately occurs at the interface between two objects of different classes.
The borders become fuzzy and details fade away.
However, on examples from \voc, at very high compression, the boat disappears and a disproportional amount of misclassification occurs.

\end{document}